\documentclass[conference]{IEEEtran}
\IEEEoverridecommandlockouts
\usepackage{cite}
\usepackage{amsmath,amssymb,amsfonts}
\usepackage{algorithm}
\usepackage{algorithmic}
\usepackage{graphicx}
\usepackage{textcomp}
\usepackage{xcolor}

\usepackage{tcolorbox}
\usepackage{xcolor}
\usepackage{caption} 
\usepackage{multirow}
\usepackage{booktabs}
\usepackage{graphicx}
\usepackage{lscape}
\usepackage{subcaption}
\usepackage{url}

\def\BibTeX{{\rm B\kern-.05em{\sc i\kern-.025em b}\kern-.08em
    T\kern-.1667em\lower.7ex\hbox{E}\kern-.125emX}}
\begin{document}

\title{Neural Algorithmic Reasoners informed Large Language Model for Multi-Agent Path Finding

\thanks{*Corresponding author: Wenjun Wu (Email: wwj09315@buaa.edu.cn). This work was supported by the National Science and Technology Major Project (No. 2022ZD0116401)}
}


\author{
    \IEEEauthorblockN{Pu Feng\textsuperscript{1}, Size Wang\textsuperscript{2}, Yuhong Cao\textsuperscript{4}, Junkang Liang\textsuperscript{2}, Rongye Shi\textsuperscript{1,2,3}, *Wenjun Wu\textsuperscript{1,2,3}}
    \IEEEauthorblockA{\textsuperscript{1}State Key Laboratory of Complex \& Critical Software Environment, Beihang University, Beijing, China}
    \IEEEauthorblockA{\textsuperscript{2}School of Artificial Intelligence, Beihang University, Beijing, China}
    \IEEEauthorblockA{\textsuperscript{3}Hangzhou International Innovation Institute, Hangzhou, China}
    \IEEEauthorblockA{\textsuperscript{4}Department of Mechanical Engineering, National University of Singapore, Singapore}
}


\maketitle

\begin{abstract}
The development and application of large language models (LLM) have demonstrated that foundational models can be utilized to solve a wide array of tasks. However, their performance in multi-agent path finding (MAPF) tasks has been less than satisfactory, with only a few studies exploring this area. MAPF is a complex problem requiring both planning and multi-agent coordination. To improve the performance of LLM in MAPF tasks, we propose a novel framework, LLM-NAR, which leverages neural algorithmic reasoners (NAR) to inform LLM for MAPF. LLM-NAR consists of three key components: an LLM for MAPF, a pre-trained graph neural network-based NAR, and a cross-attention mechanism. This is the first work to propose using a neural algorithmic reasoner to integrate GNNs with the map information for MAPF, thereby guiding LLM to achieve superior performance. LLM-NAR can be easily adapted to various LLM models. Both simulation and real-world experiments demonstrate that our method significantly outperforms existing LLM-based approaches in solving MAPF problems.
\end{abstract}


\section{Introduction}
In recent years, large language models (LLM) have rapidly advanced and proven highly effective in addressing a variety of complex tasks. Their success is largely attributed to extensive pre-training that embeds a wide range of knowledge, making them highly valuable across numerous applications. Meanwhile, integrating LLM with other computational techniques to solve specific problems has attracted considerable interest~\cite{zhu2024knowagent}.

In multi-agent systems, where coordination and cooperation are crucial, LLM has been increasingly used to enhance inter-agent collaboration, improving its ability to tackle complex tasks. While research has focused on agent interactions, communication, and world simulation~\cite{guo2024large}, studies on multi-agent planning remain limited.

This paper focuses on the multi-agent path finding (MAPF) problem, where multiple agents must navigate from start to goal locations without collisions. MAPF is vital in real-world applications like warehouse management and swarm control~\cite{surynek2022problem}. Traditional approaches include classical methods like Conflict-Based Search (CBS)~\cite{sharon2015conflict}, as well as learning-based strategies using reinforcement learning~\cite{sartoretti2019primal,li2023mixture, feng2023mact,yu2024leveraging} and graph neural networks~\cite{li2020graph}. Leveraging LLM for MAPF offers a way to mitigate the slow training and high computational demands of current methods. However, existing work~\cite{chen2024solving} on integrating LLM into MAPF is limited and does not fully explore their potential due to challenges in understanding spatial constraints and collaborative strategy formulation—areas where LLMs still fall short.

\begin{figure}[t]
  \centering
  \includegraphics[scale=0.35]{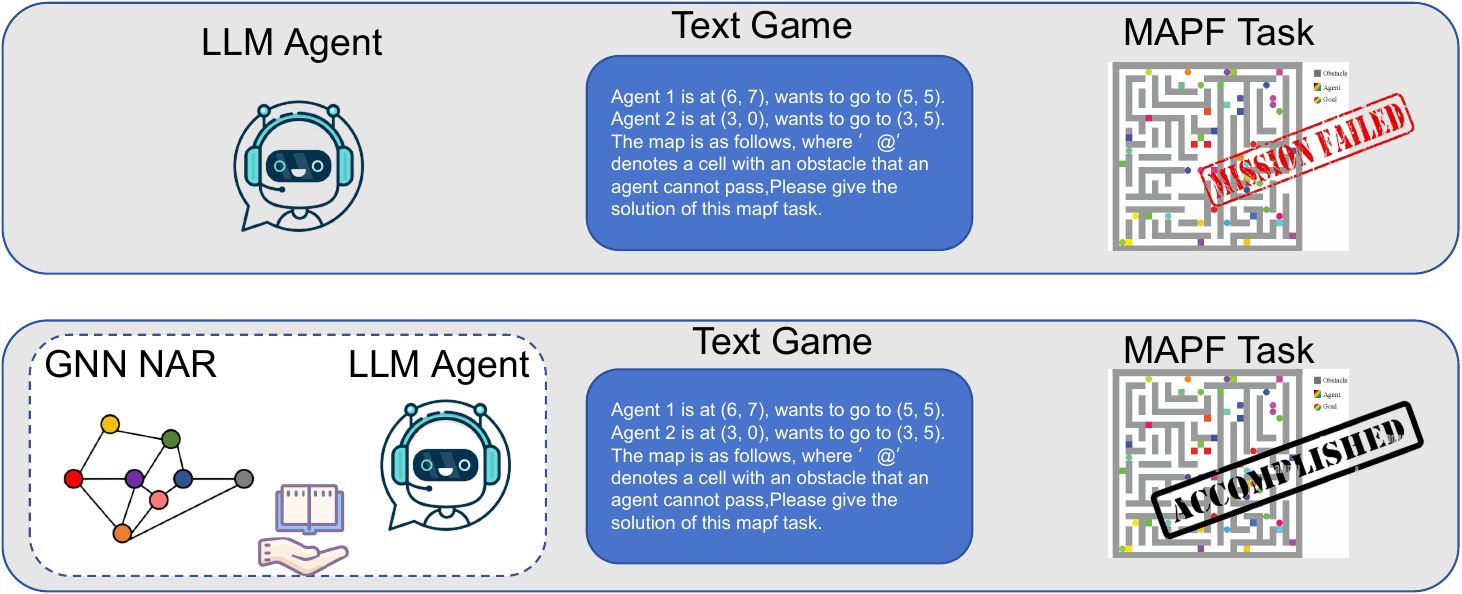}
  \caption{Limitations of existing LLM for MAPF. Informed with GNN-based NAR, our method performs better.}
  \label{intro}
  \vspace{-0.25in}
\end{figure}

To improve the performance of LLM in MAPF tasks, we propose a novel framework, LLM-NAR, which leverages neural algorithmic reasoners (NAR) to enhance LLM's ability to process spatial map information for MAPF. The framework integrates three key components: an LLM for MAPF, a GNN-based NAR, and a cross-attention mechanism.

The first component, LLM for MAPF, employs a tailored prompt interaction strategy specifically designed for MAPF tasks. Scenario-specific information is fed into LLM to generate directives for each agent at each timestep. We periodically update the LLM’s understanding of the map's state to maintain the accuracy and relevance of its outputs. A key feature of our approach is the reset mechanism, which helps keep the LLM "aware" and prevents information loss. The second component, GNN-based NAR, builds a graphical representation that captures the map's intricacies and the spatial relationships between agents. This graphical model distills critical spatial and relational insights, which are essential for effective path planning. The third component, the cross-attention mechanism, fuses the token outputs from the LLM with the graph representation produced by the GNN-based NAR. This fusion enhances contextual understanding by aligning linguistic instructions with spatial data. Finally, we train the cross-attention network by minimizing the loss between the action output from the final layer and the expert strategy derived from CBS~\cite{sharon2015conflict}.

Our approach leverages a GNN-based NAR to enhance the LLM's understanding of map information for the MAPF task, requiring only a few thousand training steps to train the cross-attention mechanism. In contrast to the hundreds of thousands or even millions of steps typically required by other learning-based methods, our approach is significantly more efficient, demanding far fewer training steps. LLM-NAR can also be easily adapted to various large language models. The key contributions of this paper are as follows:

\begin{enumerate} 
    \item We propose a novel LLM prompt mechanism tailored specifically for MAPF tasks. 
    \item We develop a GNN-based NAR to effectively represent map information. 
    \item We introduce the LLM-NAR framework, which integrates LLM and GNN-based NAR outputs via a cross-attention mechanism. 
    \item We release an open-source dataset\footnote{https://github.com/fpgod/LLM-NAR} to facilitate LLM-NAR interactions in MAPF tasks, including tokens for LLM interaction and NAR training data.
    \item We demonstrate the effectiveness of our approach through both simulation and real-world experiments. 
\end{enumerate}


\section{Related Work}
\subsection{Multi-agent Path Finding}
Multi-agent path finding~\cite{stern2019multi} tasks are classic group collaboration problems, and a significant amount of research has been conducted in this area. Current MAPF methods are mainly divided into two categories. The first category comprises traditional planning algorithms, which compute optimal or sub-optimal routes by designing specific rules; this category is represented by Conflict-Based Search (CBS)~\cite{sharon2015conflict} and its extended versions~\cite{andreychuk2021improving}. The second category consists of learning-based methods, primarily including reinforcement learning~\cite{sartoretti2019primal,feng2024safe,feng2024hierarchical}, graph neural networks~\cite{li2020graph} and transformers~\cite{wang2023scrimp}. Currently, there is relatively little research on applying large language models (LLM) to MAPF. Notably, some studies~\cite{chen2024solving} have identified existing challenges of LLM in solving MAPF problems, highlighting this as a very valuable research direction.
\subsection{Large Language Model for Multi-agent Cooperation}
Large Language Models (LLM) have begun to influence the field of multi-agent systems by enhancing communication, coordination, and decision-making among agents~\cite{ni2024tree,ni2025shieldlearner}. Leveraging their advanced natural language understanding and generation capabilities, LLM enables agents to interpret complex instructions and interact more naturally with humans and other agents~\cite{zhang2024llm}. LLM has also been used to construct interactive environments for multi-agent tasks and explore social collaboration patterns through dialogue and communication between agents~\cite{pan2024agentcoord}. These methods represent initial explorations of LLM in the multi-agent domain. Our paper goes a step further by combining LLM with graph neural networks to enhance cooperative capabilities in Multi-agent Path Finding.


\subsection{Neural algorithmic reasoning}
Neural Algorithmic Reasoning (NAR) is, in general terms, the art of building neural networks capable of capturing algorithmic computation \cite{velivckovic2021neural}, bridging the gap between precise but inflexible classical algorithms and adaptable yet less interpretable neural networks. By embedding algorithmic reasoning within neural architectures, NAR enables neural networks to perform tasks requiring logical deduction, planning, and problem-solving, enhancing their generalization and reasoning capabilities \cite{xu2018powerful}. Graph Neural Networks (GNNs) are a natural fit for NAR due to their effectiveness in representing and processing graph-structured data \cite{hechtlinger2017generalization}. Recent research~\cite{bounsi2024transformers} has explored the use of GNN-based NARs to enhance transformers in solving reasoning tasks. GNNs can also approximate classical algorithms operating on graphs, such as shortest path computation and optimization problems \cite{cappart2023combinatorial}, making them well-suited for complex tasks like MAPF, where agent relationships and environments are naturally represented as graphs \cite{ma2022graph}.

\section{Problem Formulation}
MAPF tasks involve a set of agents navigating from their initial locations to designated destinations without collisions, while minimizing travel time. Consider a set of $N$ agents, represented as $v = \{v_1, v_2, \ldots, v_N\}$, operating on an undirected graph $G = (V, E)$, where $V$ is the set of vertices and $E$ represents the edges connecting these vertices. Each agent starts at an initial vertex $s_i \in V$ and must reach a unique destination vertex $d_i \in V$. The sets of start and destination vertices are denoted by $S$ and $D$, respectively. At each discrete time step $t$, agents can either move to an adjacent vertex or remain stationary, as represented by the action set $u^t$. Collisions occur if two agents attempt to occupy the same vertex simultaneously or if an agent moves into a vertex occupied by an obstacle. The joint action set $U^t$ formed by the actions of all agents at time step $t$, is valid only if it satisfies the following conditions: (1) no two agents occupy the same vertex at any given time step ($u^t_i \neq u^t_j$ for any pair of agents $i$ and $j$), and (2) no pair of agents swap vertices within a single time step $\left(u^{t+1}_i = u^t_j \Longleftrightarrow u^{t+1}_j \neq u^t_i\right)$. The goal is to determine a sequence of valid joint action sets that allows all agents to move from their initial positions in $S$ to their destinations in $D$ in the minimum time steps.


\section{Method}

\begin{figure*}[h]
  \centering
  \includegraphics[scale=0.5]{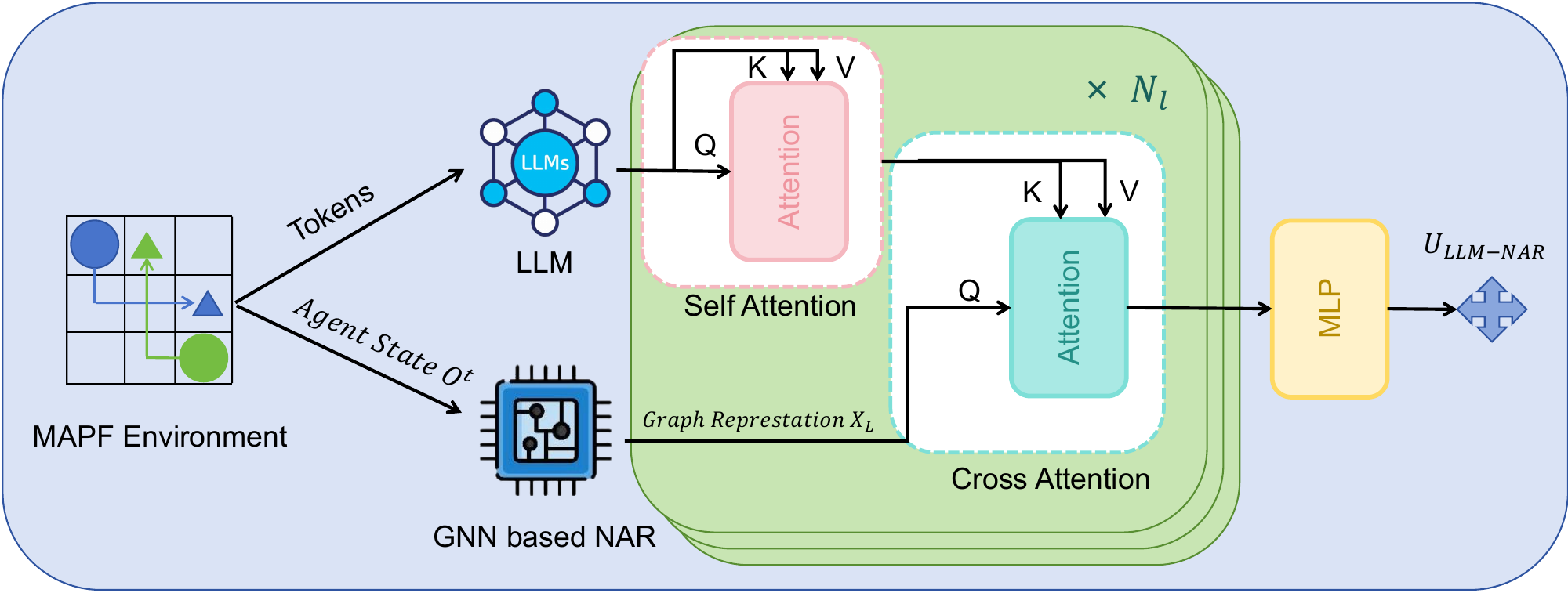}
  \caption{Framework of LLM-NAR, which consists of LLM for MAPF, GNN-based NAR and cross-attention.}
  \label{framework}
  \vspace{-0.2in}
\end{figure*}

The LLM-NAR framework is designed to address MAPF challenges by integrating three key components: LLM for MAPF, GNN-based NAR, and a cross-attention mechanism. First, we construct a novel prompt framework for LLM to handle MAPF tasks. Then, the GNN-based NAR plays a crucial role in understanding the spatial structure of the map, capturing essential relationships between agents and their environment. We leverage the capabilities of the GNN-based NAR through a cross-attention mechanism, allowing the LLM to enhance its performance. Fig.~\ref{framework} illustrates the architecture of the LLM-NAR framework. 

\begin{algorithm}[h]
    \caption{Neural Algorithmic Reasoners informed Large Language Model for Multi-Agent Path Finding}
    \label{alg1}
    \begin{algorithmic}[1] 
        \STATE Initialize MAPF task map set $\mathcal{M}$, paprameter of GNN-NAR network.
        
        \FOR{each MAPF task in $\mathcal{M}$}
            \STATE Use CBS to obtain optimal actions $\{(U^t)^*\}$ for the task.
            \STATE Train GNN-NAR network based on the obtained $\{(U^t)^*\}$ and optimize it according to Eq.~\ref{gnn loss}.
        \ENDFOR
        
        \STATE Initialize cross-attention network, max training steps, and episode length T.

        \FOR{each episode}
            \FOR{t=1 to T}
                \STATE Based on the current observation $O^t$, get GNN representation $X_L$.
                \STATE Based on the current observation $O^t$, get LLM output $\Theta^t$.
                \STATE Combine $X_L$ and $\Theta^t$ into the cross-attention mechanism to get the LLM-NAR action $U^t_{\text{LLM-NAR}}$.
                \STATE Update cross-attention network according to Eq.~\ref{llm-nar loss}.
            \ENDFOR
        \ENDFOR
    \end{algorithmic}
\end{algorithm}

\subsection{Overview of LLM-NAR}

In our algorithmic framework, we integrate three core components: GNN-NAR, LLM, and a cross-attention mechanism that synergizes the strengths of these two. The pseudocode for our method is presented in Algorithm~\ref{alg1}. 

CBS, as an optimal algorithm, has the drawback of high computational cost but offers the advantage of providing optimal solutions, making it suitable for use as training data labels. We first employ CBS~\cite{sharon2015conflict} to generate optimal path data for each MAPF task and use this data to pretrain the GNN-NAR network, resulting in a reasonably accurate GNN-NAR model. At this stage, we assume that GNN-NAR has acquired the capability to represent MAPF map information, denoted as $X_L$. Concurrently, based on the state information at each time step, we design a novel prompt format tailored specifically for MAPF tasks, which allows the LLM to generate token outputs.

By feeding both $X_L$ and the LLM-generated token outputs $\Theta^t$ into the cross-attention mechanism, we derive the final action output of LLM-NAR, $U^t_{\text{LLM-NAR}}$. The loss is then computed between $U^t_{\text{LLM-NAR}}$ and the CBS-generated optimal action $u^t$, and this loss is used to update the cross-attention network.

Throughout this process, we leverage the GNN's ability to effectively encode map structures and the LLM's planning capabilities to address the MAPF task more comprehensively. This integration mitigates the limitations of relying solely on either GNN or LLM for solving MAPF challenges. A detailed explanation of each component is provided below.

\subsection{LLM for MAPF}

\begin{figure}[h]
  \centering
  \includegraphics[scale=0.35]{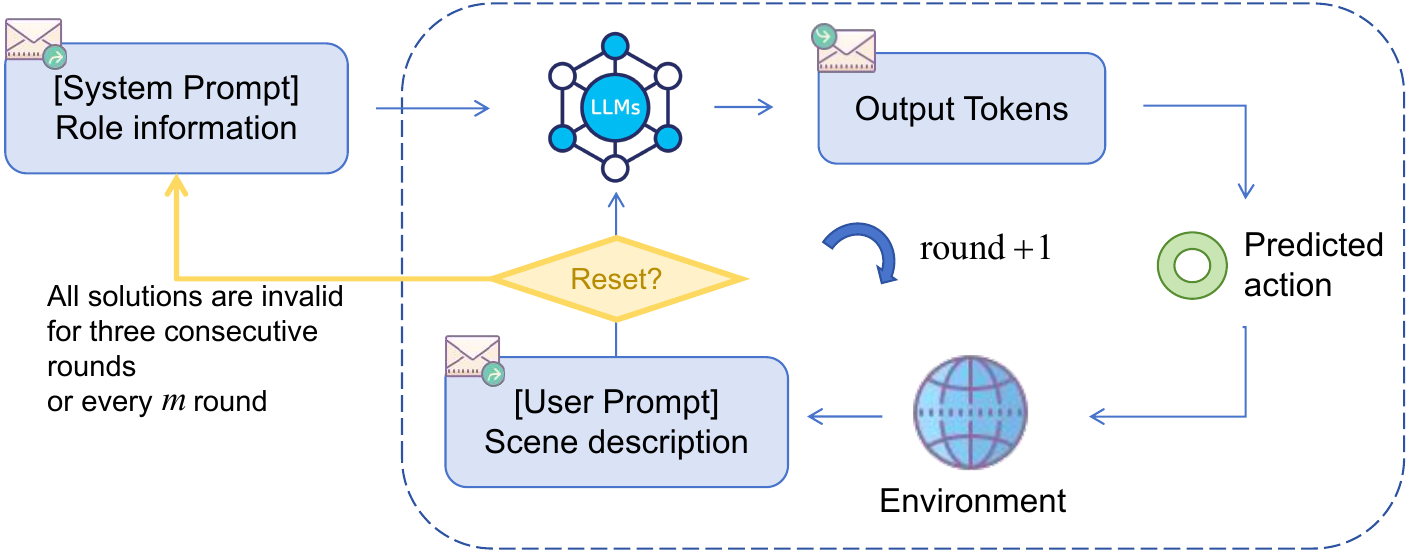}
  \caption{Workflow of LLM for MAPF}
  \label{llmworkflow}
  \vspace{-0.1in}
\end{figure}

In this module, we focus on utilizing LLM to address the MAPF problem. Inspired by~\cite{chen2024solving}, we construct our LLM framework for MAPF with several improvements, which features a specially designed scene description prompt and a reset mechanism for LLM, as shown in Fig.~\ref{llmworkflow}.

We provide LLM with a system prompt of role information, allowing it to function as the solver for MAPF problem. Based on the specific map and task conditions, we progressively provide LLM with scene descriptions step by step. This information includes the agents' current positions, target locations, a text-based map description, and the positions of both agents and obstacles, as shown in Fig.~\ref{fig:llm map prompt}. The step-by-step design breaks the entire planning task into manageable single-step tasks. 

We extract the solutions from LLM output tokens $\Theta^t$ at timestep $t$. Any answers that violate the rules or are deemed invalid are corrected to the "stay" action to ensure the smooth operation of the workflow. Then we feed the predicted actions into the environment, enabling the state transition and proceeding to the next step. Additionally, a reset mechanism is employed to modify the prompt process based on an evaluation of the LLM's performance.


\begin{figure}[h]
    \centering
    \begin{tcolorbox}[sharp corners, boxrule=0.5mm, colback=white]

        \textbf{Agent 1} is at (6, 7), wants to go to (5, 5).\\
        \textbf{Agent 2} is at (3, 0), wants to go to (3, 5).\\
        \textcolor{blue}{The map is as follows, where '@' denotes a cell with an obstacle that an agent cannot pass, and '.' denotes an empty cell that an agent can pass.}\\
        \textcolor{blue}{The lower-left cell is (0,0) and the lower-right cell is (0,7):}\\
        \textcolor{blue}{...}\\
        \textcolor{blue}{......@.}\\
        \textcolor{blue}{..@.....}\\
        \textcolor{blue}{...}\\
        \textcolor{red}{The coordinates of the obstacles: (2,5) (1,7) (6,3)... }\\
        \textcolor{red}{The coordinates of the agents: (6,7) (3,0).} 
        
    \end{tcolorbox}
    \caption{Example of the user prompt for describing the scenario. The black text describes the mission information. The blue text describes the map information. The red text is a mathematical representation of the scene.}
    \label{fig:llm map prompt}
    \vspace{-0.1in}
\end{figure}

Compared to previous approaches that employed conflict checkers to generate feasible actions prompt at each step~\cite{chen2024solving}, we simplify the workflow by integrating the conflict checker into the environment transition process and abandoned the feasible action prompt. Instead, we directly provide step-by-step scenario information as input before the LLM generates actions. We find that in later stages of problem-solving, LLM often encountered issues such as loss or confusion of scene information (e.g., forgetting target locations or providing invalid actions). Prompting with feasible actions proved insufficient to mitigate these problems. To emphasize the agents' objectives, we utilize detailed scene descriptions instead of merely feasible actions. Our approach ensures that LLM receives information prompts at each step. This assists LLM in understanding its current stage in the process and aligning their actions with the intended goals. 

Additionally, we introduce a reset mechanism for the LLM. The LLM is reset when poor performance is detected, and the task continues from the current position, treating it as the new starting point. We consider LLM needs a restart either after completing a set number of rounds (denoted as $m$, $m$=5 in our experiments) or when it generates only invalid solutions for three consecutive rounds. In such cases, we can promptly stop LLM from going down the wrong path and "clear its head" by rebooting it. The task is considered complete when the agents reach their goals or the entire planning process (including resets) reaches three times the map length, which allows sufficient room for agents to traverse the longest path and make any necessary detours.





\subsection{GNN-based NAR}

\begin{figure}[h]
  \centering
  \includegraphics[scale=0.28]{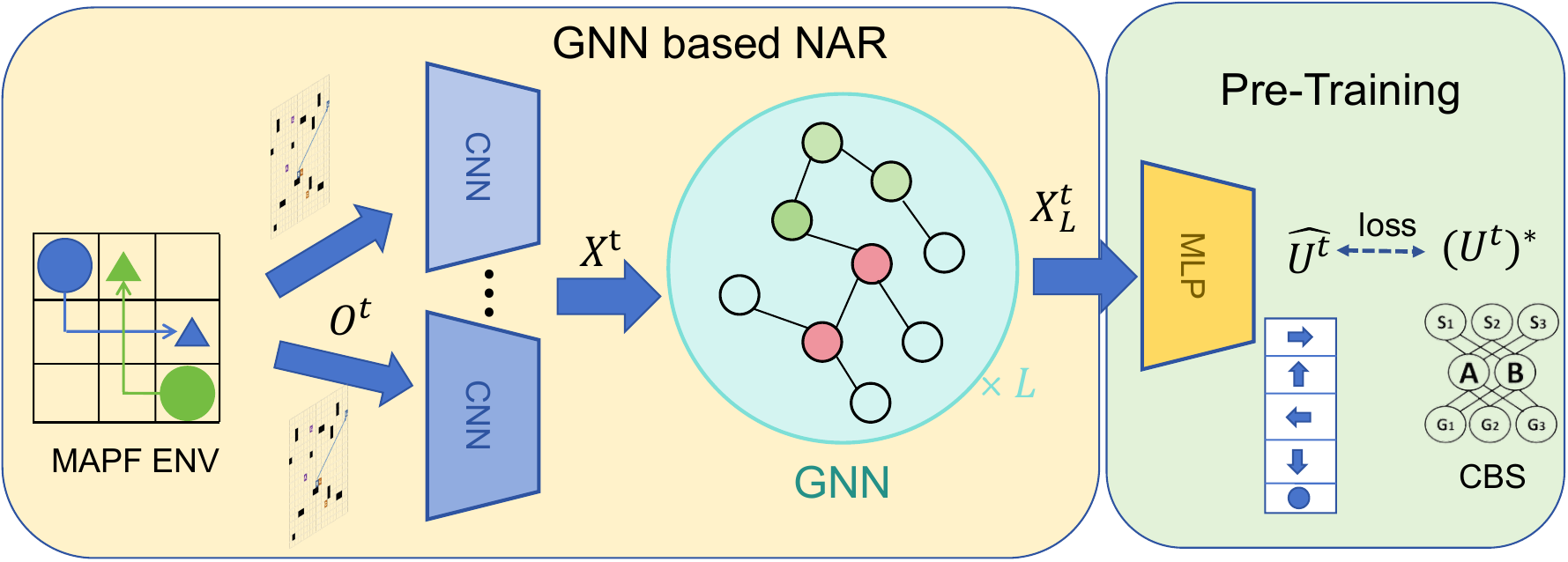}
  \caption{Pretrained GNN-based NAR}
  \label{nar}
  \vspace{-0.2in}
\end{figure}

In this section, we detail the deployment of a pre-trained GNN, configured as a Neural Algorithmic Reasoner (NAR), as depicted in Fig.~\ref{nar}, to assist the LLM in interpreting spatial and relational dynamics within the MAPF scenario. The GNN plays a critical role in enhancing the understanding of the map's structure and the proximity between agents, improving decision-making by extracting and integrating complex spatial features and state information.

Each agent, identified by an index \(i\), gathers local environmental data at each timestep \(t\), represented as \(o^t_i\). These observations are assembled into a matrix for all agents: $O^t = [o^t_1, o^t_2, \ldots, o^t_N]$, where \(N\) denotes the total number of agents. This observation matrix \(O^t\) is processed by a Convolutional Neural Network (CNN), which translates the raw data into a series of feature vectors \(X^t\):
\begin{equation}
X^t = \text{CNN}(O^t) = [x^t_1, x^t_2, \ldots, x^t_N].
\end{equation}

Please note that our ultimate goal is to enable the LLM to derive superior strategies from more effectively processed map information. Within the framework of our GNN-based NAR, we have defined a connectivity structure among agents, represented by the adjacency matrix \( C^t \), to delineate the relationships between agents. Intuitively, if an agent \( i \)'s strategy might potentially collide with neighbor agent \( j \), then the state information of \( j \) is invaluable for training the strategy of \( i \). To this end, we use a graph convolution to aggregate observational data using \( C^t \), expressed through the operation \( \mathcal{A}(X^t; C^t) \):
\begin{equation}
\mathcal{A}(X^t; C^t) = [C^t X^t]_{i} = \sum_{j=1}^N [C^t]_{ij} [X^t]_{j} = \sum_{j: v_j \in \mathcal{N}_i} c^t_{i} x^t_{j}.
\end{equation}


The GNN architecture consists of \(L\) layers, with each layer updating its state by integrating outputs from the previous layer through the defined graph convolution:
\begin{equation}
X_{\ell} = \sigma[\mathcal{A}_{\ell}(X_{\ell-1}; C)] \quad \text{for} \quad \ell = 1, \ldots, L,
\end{equation}
where \(\sigma\) represents a non-linear activation function that enhances the model’s capacity to discern complex patterns. After processing through the GNN layers, the resulting features are fed into a Multi-Layer Perceptron (MLP) to generate the predicted actions $\hat{U}^t = \text{MLP}(X_L^t)$.


The optimization process for the GNN is aimed at minimizing the discrepancies between the predicted actions $\hat{U}^t$ and the optimal actions $(U^t)^*$, derived from CBS~\cite{sharon2015conflict} as the expert policy. The objective is to minimize the loss, which can be formally expressed as:
\begin{equation}
\min_{\mathrm{CNN}, \{\mathrm{A}_{\ell}\}, \mathrm{MLP}} \sum_{(\{U^t\}, \{\mathbf{O}^t_i\}) \in \mathcal{T}} \sum_t \mathcal{L}((U^t)^*, \hat{U}^t).
\label{gnn loss}
\end{equation}
Here, \(\mathcal{L}\) denotes the loss function, which quantifies the fidelity of the GNN's predictions in replicating the expert policy's actions. The set \(\mathcal{T}\) represents the training dataset, consisting of sequences of observations and corresponding actions.

By optimizing the GNN to closely align with the expert policy, we enhance the accuracy of the graph representations, which in turn supports more effective information processing. Through this process, we obtain a pre-trained GNN-based NAR that provides knowledge-rich spatial information for the subsequent cross-attention process in the LLM-NAR framework, facilitating superior decision-making.





\subsection{Cross-Attention}
Inspired by the Flamingo framework \cite{bounsi2024transformers} in multi-modal tasks, our approach utilizes a multi-layer cross-attention mechanism to fuse diverse modalities of information. This section describes the implementation of our cross-attention mechanism, which is pivotal in synthesizing the final policy by integrating outputs from both LLM and the GNN-based NAR. Within our method, both Self-Attention and Gated Cross-Attention blocks are utilized to refine the integration of linguistic and spatial data. Specifically, we obtain the LLM output token $\Theta^t$, alongside the graph representation $X_{L}^t$ derived from GNN-based NAR. The token $\Theta^t$ is first processed through a Self-Attention block and further enhanced via a residual connection. Subsequently, these tokens are subjected to cross-attention with the graph representations $X_{L}^{t}$ through the Gated Cross-Attention block. 

The final output of the LLM-NAR interaction at the subsequent timestep is formulated as:
\begin{equation}
\mathbf{T}^{t}=\operatorname{FFN}\left(\operatorname{softmax}\left(\frac{\left(\mathbf{\Theta}^{t} \mathbf{Q}^t\right)^{\top} \mathbf{X}_L^{t} \mathbf{K}^t}{\sqrt{d_k}}\right) \mathbf{X}_L^{t} \mathbf{V}^t\right)
\label{cross}
\end{equation}

In this equation, \(\mathbf{Q}^t, \mathbf{K}^t \in \mathbb{R}^{k \times d_k}, \mathbf{V}^t \in \mathbb{R}^{k \times k}\) represent the query, key, and value transformations used in the cross-attention mechanism, respectively. FNN is a feed-forward network. Notably, the queries are derived from the LLM tokens, while the keys and values are obtained from the graph representations. To maintain the LLM’s intrinsic knowledge during the initial phase of training, we initially close the gate of the cross-attention mechanism, gradually introducing more complex interactions as training progresses.

\begin{table*}[htbp]
\centering
\renewcommand{\arraystretch}{1.2} 
\caption{Success Rate of different method}
\label{success rate}

\resizebox{\textwidth}{!}{
\begin{tabular}{c||ccccc||ccccc||ccccc}
\toprule
  \multirow{2}{*}{Model}      & \multicolumn{5}{c||}{8x8 empty map with 2,4,8,10,16 agents}    & \multicolumn{5}{c||}{20x20 empty map with 2,4,8,10,16 agents}     & \multicolumn{5}{c}{28x28 empty map with 2,4,8,10,16 agents}       \\ 
    & 2 & 4 & 8 & 10 & {16} & 2 & 4 & 8 & 10 & {16} & 2 & 4 & 8 & 10 & 16 \\ \midrule 
Qwen2 & 55.00\% & 50.00\% & 30.00\% & 18.75\% & {9.38\%} & 50.00\% & 25.00\% & 10.00\%& 9.38\%  & {4.69\%} & 37.50\% & 18.75\% & 7.50\% & 4.69\% & 3.13\%  \\
Gemma2 & 100.00\% & 93.75\% & 68.75\% & 55.00\% & {43.75\%} & 100.00\% & 93.75\% & 53.13\%& {46.88\%} & 27.50\%  & 100.00\% & 87.50\% & 40.63\% & 40.00\% & 28.13\% \\
LLaMA3 & 100.00\% & 81.25\% & 71.25\% & 45.00\% & {31.13\%} & 93.75\% & 56.25\% & 50.00\% & 30.00\% & {21.88\%} & 100.00\% & 75.00\% & 60.00\% & 56.25\% & 26.56\% \\
GPT-3.5-turbo& 100.00\% & 90.63\% & 75.00\% & {59.38\%}& 50.00\%  & 75.00\%  & 71.88\% & 65.00\% & 62.50\%& {56.94\%}& 100.00\% & 75.00\% & 62.50\% & 56.25\% & 46.88\% \\ \midrule
LLM-NAR & \textbf{100.00\%} & \textbf{100.00\%} & \textbf{93.75\%} & \textbf{75.00\%} & {\textbf{67.19\%}} & \textbf{100.00\%} & \textbf{93.75\%} & \textbf{93.75\%} & \textbf{80.00\%} & {\textbf{65.63\%}} & \textbf{100.00\%} & \textbf{100.00\%} & \textbf{75.00\%} & \textbf{70.00\%} & \textbf{60.94\%} \\ \midrule \midrule 
       & \multicolumn{5}{c||}{8x8 map (10\% obstacles) with 2,4,8,10,16 agents}    & \multicolumn{5}{c||}{16x16 map (10\% obstacles) with 2,4,8,10,16 agents}     & \multicolumn{5}{c}{20x20 map (10\% obstacles) with 2,4,8,10,16 agents}       \\ 
    & 2 & 4 & 8 & 10 & {16} & 2 & 4 & 8 & 10 & {16} & 2 & 4 & 8 & 10 & 16 \\ \midrule 
Qwen2  &  37.50\% & {23.44\%} & 15.00\% & 12.50\% & 6.25\% &   25.00\% & 12.50\% & 6.25\% & 6.25\%  & {3.13\%}& 25.00\%& 12.50\%  & 9.38\% & 4.69\%& 2.50\%  \\
Gemma2 & 75.60\% & 62.50\%& {54.69\%}&  37.50\%  & 20.00\%  & 75.00\% & {42.19\%}& 40.63\%& 22.50\%&  12.50\%    & 75.00\% & 43.75\% & 31.25\%& 20.00\%& 18.75\%   \\
LLaMA3 & 81.25\%&  62.50\%  & 52.50\%& 46.88\%  & {39.06\%} & 37.50\%& 25.00\%&  17.50\% &  12.50\%   & {7.81\%}& 75.00\% & 50.00\% & 17.50\%& 12.50\%  & 7.81\% \\
GPT-3.5-turbo  & 81.25\%& 75.00\%  & 50.00\% & 42.50\% & {36.06\%}  & 62.50\% & 46.88\% & 50.00\% & {50.00\%}& 37.50\%& 75.00\%  & 71.88\% & 62.50\%& 60.00\% & 48.44\% \\ \midrule
LLM-NAR &  \textbf{100.00\%}& \textbf{100.00\%} & \textbf{78.13\%} & \textbf{72.50\%} & {\textbf{65.75\%}} &  \textbf{87.50\%} & \textbf{75.00\%} & \textbf{69.38\%} & \textbf{64.50\%} & {\textbf{60.25\%}}& \textbf{87.50\%} & \textbf{81.25\%} & \textbf{75.00\%} & \textbf{65.00\%} & \textbf{59.13\%} \\ \bottomrule

\end{tabular}}
\end{table*}

\begin{table*}[htbp]
\centering
\renewcommand{\arraystretch}{1.2} 
\caption{Average Step of different method}
\label{average step}

\resizebox{\textwidth}{!}{
\begin{tabular}{c||ccccc||ccccc||ccccc}
\toprule
  \multirow{2}{*}{Model}    & \multicolumn{5}{c||}{8x8 empty map with 2,4,8,10,16 agents}    & \multicolumn{5}{c||}{20x20 empty map with 2,4,8,10,16 agents}     & \multicolumn{5}{c}{28x28 empty map with 2,4,8,10,16 agents}       \\ 
    & 2 & 4 & 8 & 10 & {16} & 2 & 4 & 8 & 10 & {16} & 2 & 4 & 8 & 10 & 16 \\ \midrule 
Qwen2  & \hspace{1.5mm}0.70\hspace{1.5mm} & \hspace{1.5mm}0.81\hspace{1.5mm} & \hspace{1.5mm}0.85\hspace{1.5mm} & \hspace{1.5mm}0.88\hspace{1.5mm} & \hspace{1.5mm}{0.97}\hspace{1.5mm} & \hspace{1.5mm}0.86\hspace{1.5mm} & \hspace{1.5mm}0.90\hspace{1.5mm} & \hspace{1.5mm}0.92\hspace{1.5mm} & \hspace{1.5mm}0.98\hspace{1.5mm} &\hspace{1.5mm}1.00\hspace{1.5mm} & \hspace{1.5mm}0.79\hspace{1.5mm} & \hspace{1.5mm}1.00\hspace{1.5mm} & \hspace{1.5mm}1.00\hspace{1.5mm} & \hspace{1.5mm}1.00\hspace{1.5mm} & \hspace{1.5mm}1.00\hspace{1.5mm} \\
Gemma2 & 0.25 & 0.46 & 0.65 & 0.75 & {0.90} & 0.51 & 0.53 & 0.79 & 0.88 & {0.93} & 0.52 & 0.67 & 0.80 & 0.83 & 0.94 \\
LLaMA3 & 0.31 & 0.44 & 0.52 & 0.72 & {0.89} & 0.54 & 0.63 & 0.75 & 0.90 & {0.95} & 0.49 & 0.55 & 0.64 & 0.83 & 0.88 \\
GPT-3.5-turbo& 0.36 & 0.49 & 0.59 & 0.65 & {0.83} & 0.51 & 0.55 & 0.64 & 0.69 & {0.79}& 0.35 & 0.55 & 0.61 & 0.63 & 0.70 \\ \midrule
LLM-NAR & \textbf{0.22} & \textbf{0.29} & \textbf{0.48} & \textbf{0.47} & {\textbf{0.63}} & \textbf{0.31} & \textbf{0.39} & \textbf{0.47} & \textbf{0.54} & {\textbf{0.64}} & \textbf{0.23} & \textbf{0.45} & \textbf{0.48} & \textbf{0.50} & \textbf{0.54} \\ \midrule \midrule 
       & \multicolumn{5}{c||}{8x8 map (10\% obstacles) with 2,4,8,10,16 agents}    &  \multicolumn{5}{c||}{16x16 map (10\% obstacles) with 2,4,8,10,16 agents}     & \multicolumn{5}{c}{20x20 map (10\% obstacles) with 2,4,8,10,16 agents}       \\ 
    & 2 & 4 & 8 & 10 & 16 & 2 & 4 & 8 & 10 & {16} & 2 & 4 & 8 & 10 & 16 \\ \midrule 
Qwen2  &  0.88 & 1.00 & 1.00 & 1.00 & {1.00} &  0.90 & 0.93 & 0.95 & 1.00 & {1.00} & 0.95 & 1.00 & 1.00 & 1.00 & 1.00 \\
Gemma2 &  0.50 & 0.56 & 0.70 & 0.76 & {0.91} &  0.61 & 0.70 & 0.81 & 0.89 &{0.93} & 0.58 & 0.71 & 0.84 & 0.92 & 0.98 \\
LLaMA3 &  0.63 & 0.66 & 0.75 & 0.98 & {1.00} &  0.75 & 0.79 & 0.84 & 0.94 & {1.00} & 0.51 & 0.74 & 0.98 & 1.00 & 1.00 \\
GPT-3.5-turbo &  0.56 & 0.59 & 0.80 & 0.87 & {0.88} &  0.56 & 0.57 & 0.74 & 0.78 & {0.81} & 0.45 & 0.68 & 0.71 & 0.78 & 0.84 \\ \midrule
LLM-NAR &  \textbf{0.31} & \textbf{0.34} & \textbf{0.51} & \textbf{0.60} & {\textbf{0.62}} &  \textbf{0.32} & \textbf{0.52} & \textbf{0.60} & \textbf{0.65} & {\textbf{0.72}} & \textbf{0.33} & \textbf{0.55} & \textbf{0.61} & \textbf{0.63} & \textbf{0.71} \\

\bottomrule

\end{tabular}}
\end{table*}

This process repeats \(N_l\) times, which is set to three in this paper. As we enter the next round of cross-attention, the previous \(T^t\) is used as the input \(\Theta^t\) for the next layer, and a new \(T^t\) is generated according to Eq.~\ref{cross}. Subsequently, the predicted actions are extracted from the final layer output \(T^t_{N_l}\). During the training of the LLM-NAR, we freeze the parameters of both the LLM and the pre-trained GNN-based NAR, focusing only on updating the cross-attention component. The optimization process is similar to that used during the pre-training of the GNN. We train the model by minimizing the discrepancy between the actions output of the LLM-NAR, denoted \(U_{LLM-NAR}\), and the optimal actions \((U^t)^*\) derived from CBS~\cite{sharon2015conflict}:
\begin{equation}
\min \mathcal{L}(U_{LLM-NAR}^t, (U^t)^*),
\label{llm-nar loss}
\end{equation}
By integrating an LLM with a GNN-based NAR in the LLM-NAR framework, we enhance the LLM's ability to comprehend map information. This approach is well-suited for MAPF tasks at various scales due to the scalability of both the LLM and GNN. In practice, only the cross-attention component requires training. LLMs, such as GPT or LLaMA, can function locally or via an API, while the GNN, as detailed in \cite{li2020graph}, can be utilized through pre-training, streamlining deployment in real-world applications. Notably, training the cross-attention mechanism requires only a small number of samples and steps, significantly fewer than methods like reinforcement learning.


\section{Experiment Results}
This section demonstrates our LLM-NAR's superiority via experiments in both simulation and real-world MAPF tasks.

\subsection{Experiment Setup}

To validate the effectiveness of our approach, we conducted experiments on maps with varying sizes and obstacle densities. Given the specific conditions of each task and potential performance variations of the LLM, we ran each algorithm 10 times on every map and averaged the results to obtain our experimental data. Notably, we use 100 execution cases to train the cross-attention mechanism and complete the process in just 5K training steps.

We employed two evaluation metrics to assess performance: \textit{success rate} and \textit{average step}. The \textit{success rate} (\(R\)) is
the percentage of agents that reach their goals in all episodes, expressed as $R = n_{\text{success}}/{n}$, where \(n_{\text{success}}\) represents the number of agents who succeed, and \(n\) is the total number of agents. The \textit{average step} (\(\delta\)) is calculated as the mean executed path length for all agents, normalized by the maximum number of steps allowed in the map. It is computed by dividing the total number of steps taken by all agents by the product of the number of agents and the maximum steps: $\delta = \frac{\text{Total Steps}}{\text{Number of Agents} \times \text{Max Steps}}$.

\begin{figure}[ht]
    \centering
    \begin{subfigure}{0.48\linewidth}
        \centering
        \includegraphics[width=\textwidth]{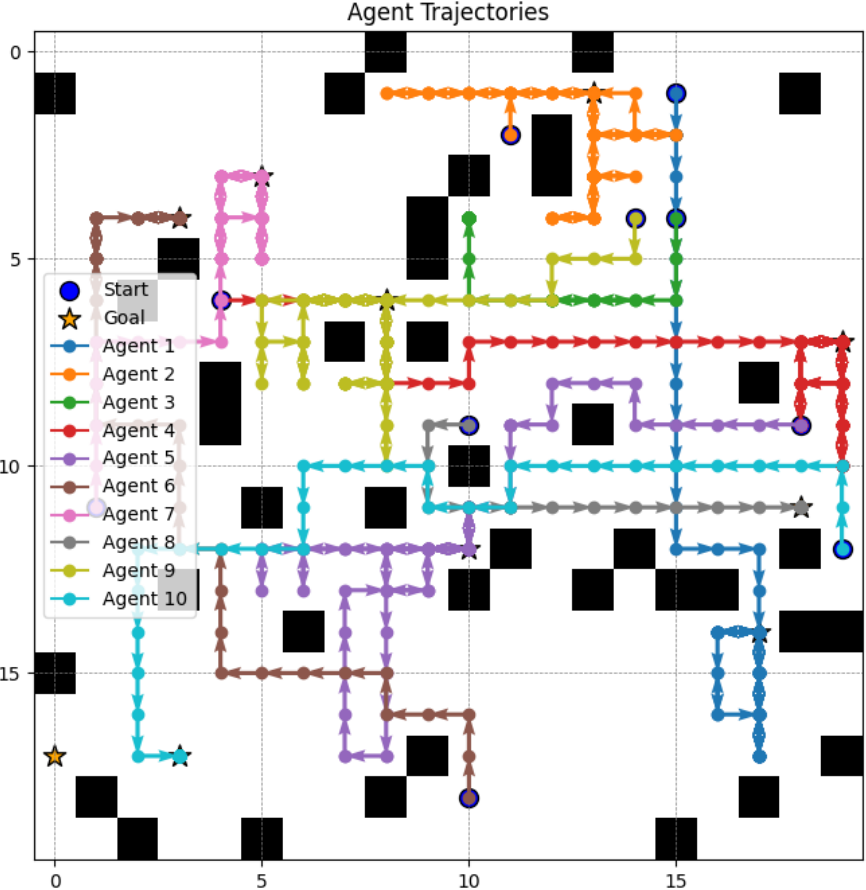}
        \caption{Trajectories of the robots with GPT.}
        \label{gpt_tra}
    \end{subfigure}
    \hfill
    \begin{subfigure}{0.48\linewidth}
        \centering
        \includegraphics[width=\textwidth]{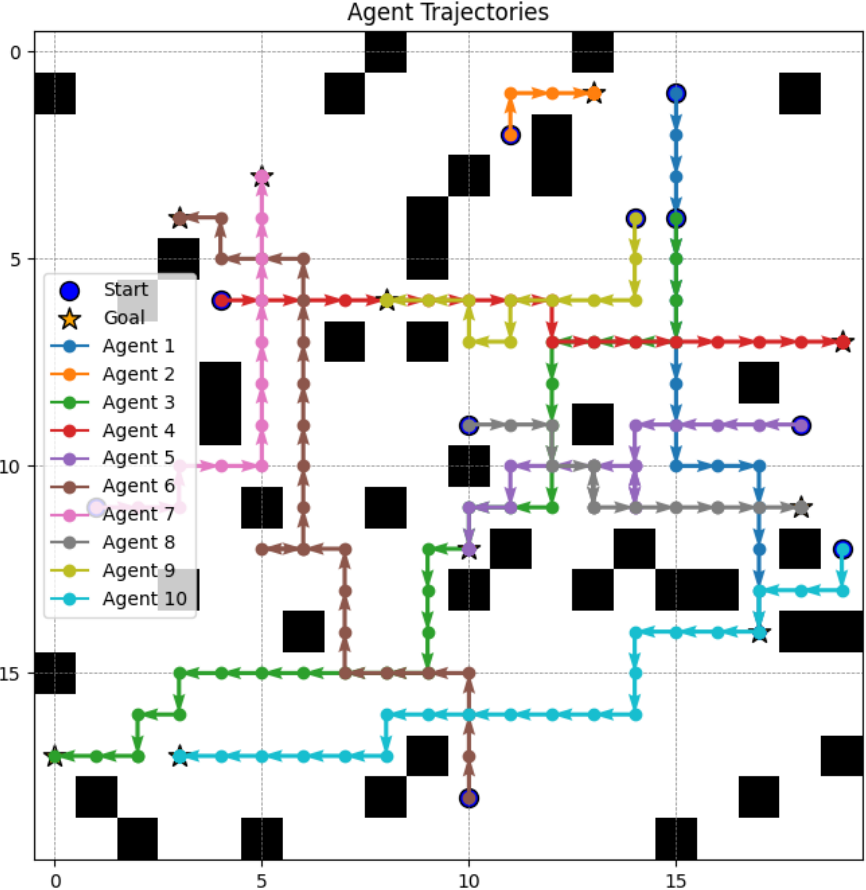}
        \caption{Trajectories of the robots with our method.}
        \label{ours_tra}
    \end{subfigure}
    \caption{Comparison of the agent trajectories in the $20 \times 20$ map with a 10\% obstacle ratio and 10 agents.}
    \label{sim_tra}
    \vspace{-0.1in}
\end{figure}

Our LLM-NAR method is compatible with various LLM models. In this study, we primarily used GPT-3.5-turbo as the foundational model, while also selecting several commonly used LLMs as baselines for comparative analysis. Due to budgetary constraints, we have not performed comprehensive testing on GPT-4. However, we believe that if our method demonstrates improvements with GPT-3.5, it should similarly enhance more advanced models in the future. The primary objective of our experiments is to showcase the practical value of our approach, highlighting its applicability and scalability to newer, more advanced models.

\begin{figure*}[ht]
    \centering
    \begin{minipage}{0.32\textwidth}
        \centering
        \includegraphics[width=\textwidth]{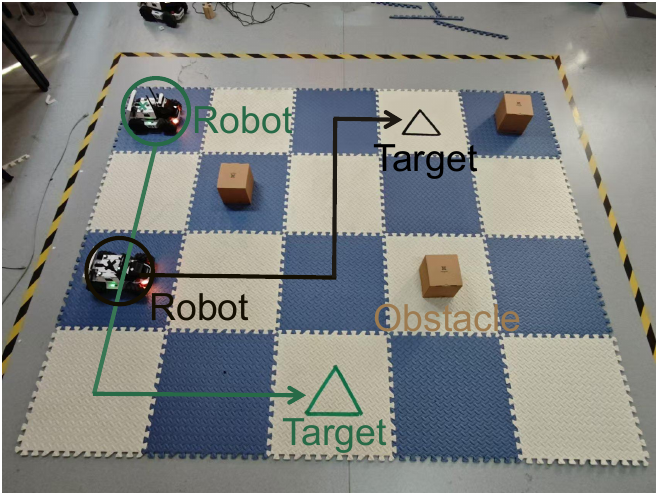}
        \subcaption{Two robots}
        \label{tra-gpt}
    \end{minipage}
    \hfill
    \begin{minipage}{0.32\textwidth}
        \centering
        \includegraphics[width=\textwidth]{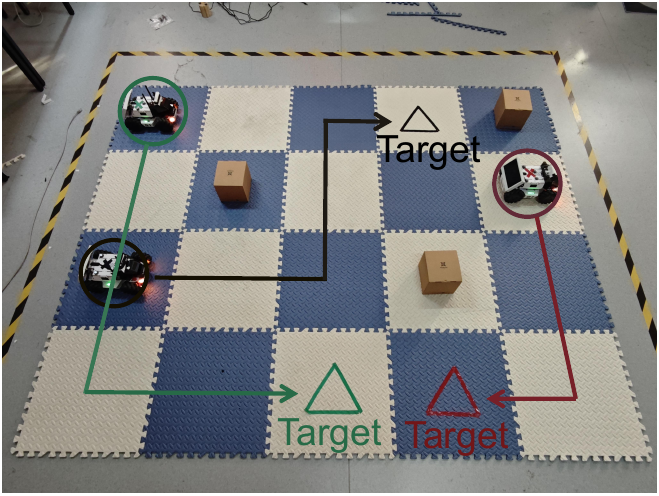}
        \subcaption{Three robots}
        \label{tra-llmnar}
    \end{minipage}
    \hfill
    \begin{minipage}{0.32\textwidth}
        \centering
        \includegraphics[width=\textwidth]{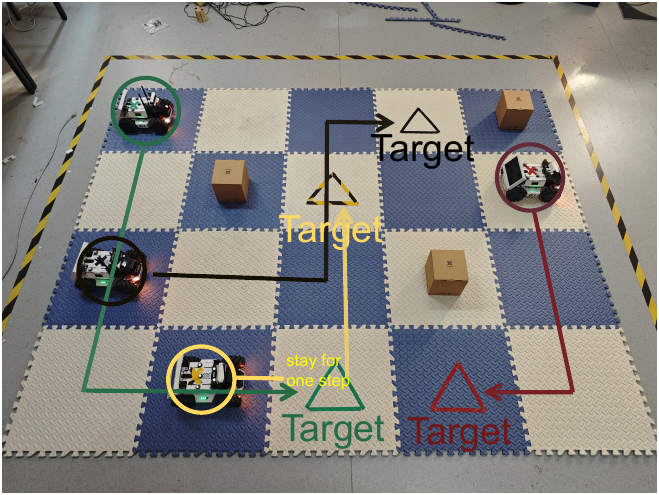}
        \subcaption{Four robots}
        \label{real3}
    \end{minipage}
    \caption{Real-world experiment with two, three, and four robots.}
    \label{real}
    \vspace{-0.3in}
\end{figure*}

\subsection{Simulation Results}
We conducted simulations to test tasks on maps of different sizes, with varying numbers of agents, and in environments both with and without obstacles. We used the \textit{success rate} metric to represent the algorithm's ability to successfully complete tasks, and the \textit{average step} to reflect the algorithm's path efficiency. The \textit{success rate} is shown in Table~\ref{success rate}, and the \textit{average step} is presented in Table~\ref{average step}.

As shown in Table~\ref{success rate}, on each map, the success rates of all methods decrease as the number of agents increases. This suggests that task difficulty increases with the number of agents. When the number of agents is small, the baseline models LLaMA3 and GPT perform well, but their success rates decline rapidly as the number of agents increases. In contrast, LLM-NAR consistently achieves higher success rates than the other baselines across different numbers of agents, and maintains good performance even when the number of agents is large. In the 20$\times$20 map without obstacles, when the number of agents is 10, GPT achieves a success rate of 62.50\%, while LLM-NAR reaches 80.00\%, significantly higher than other baselines. When the number of agents increases to 16, the success rates of the baseline models drop below 60\%, whereas LLM-NAR maintains a success rate of 65.63\%.

Furthermore, in more complex tasks involving obstacles, LLM-NAR also outperforms the other baselines. Due to the impact of obstacles, Qwen2's performance is relatively poor. LLaMA3 performs relatively well when the number of agents is small, while GPT performs better than other baselines when the number of agents is large. However, LLM-NAR consistently outperforms all baseline models across different map sizes and numbers of agents, indicating that in environments with obstacles, LLM-NAR enhances task performance.

Table~\ref{average step} presents the results for the average step metric. Overall, across tasks with different map sizes and numbers of agents, LLM-NAR requires the fewest average steps to reach the target points, resulting in shorter path lengths. In the 20$\times$20 map without obstacles, when the number of agents is 10, LLM-NAR reaches the goal using only $0.54 \times \text{Max Steps}$, while other methods require at least $0.69 \times \text{Max Steps}$. When the number of agents increases to 16, LLM-NAR's average step of $0.64$ still outperforms GPT's $0.79$. In maps with obstacles, the number of steps required by each method generally increases compared to the empty map. However, when comparing different methods, our LLM-NAR still shows significant advantages, outperforming all baseline models across all map sizes and numbers of agents.


Fig.~\ref{sim_tra} illustrates the task trajectories of GPT-3.5-turbo and LLM-NAR on a $20 \times 20$ map with a 10\% obstacle ratio and 10 agents. Both methods were tested under the same obstacle setting. From the trajectory visualization, it is evident that all agents using our proposed LLM-NAR successfully reached their respective goal positions on this map, whereas one agent in the GPT-based approach failed to reach its designated target. Additionally, the paths generated by LLM-NAR are shorter and more efficient compared to those of GPT, demonstrating superior path efficiency.

\subsection{Practical Analysis}

From the experimental results presented above, it can be observed that our proposed LLM-NAR significantly improves the performance of LLM in MAPF tasks. This improvement is evident in terms of higher success rates and more efficient paths, especially in tasks involving a larger number of agents and larger-scale maps. From this perspective, our proposed LLM-NAR framework provides a novel approach to solving MAPF problems by leveraging GNN-based information.

In addition to comparisons with LLM, we also conducted experiments comparing our method with traditional planning approaches, such as CBS, and reinforcement learning methods, including PRIMAL~\cite{sartoretti2019primal}, DHC~\cite{ma2021distributed}, and SCRIMP~\cite{wang2023scrimp}. 

One crucial metric for learning-based methods is training efficiency, which is typically measured by the number of steps required for convergence. Table~\ref{tab:training_steps} presents the training steps needed for different RL-based methods. As shown in the table, our LLM-NAR framework requires significantly fewer training steps compared to other RL-based approaches.

\begin{table}[h]
    \centering
    \caption{Required Training Steps}
    \label{tab:training_steps}
    \begin{tabular}{c|cccc}
        \hline
        \textbf{Method} & \textbf{LLM-NAR} & \textbf{PRIMAL} & \textbf{DHC} & \textbf{SCRIMP} \\ 
        \hline
        \textbf{Training Steps} & $5 \times 10^3$ & $3 \times 10^5$ & $3 \times 10^5$ & $3 \times 10^5$ \\
        \hline
    \end{tabular}
\end{table}

As shown in Table~\ref{tab:running_time_20x20}, which presents execution time as an indicator of computational cost, our method achieves significantly lower execution time compared to planning-based approaches such as CBS. Notably, as the number of agents increases, the runtime efficiency gap between LLM-NAR and CBS widens, further highlighting the scalability advantage of our approach.

\begin{table}[h]
    \centering
    \caption{Running Time in 20×20 Map (seconds)}
    \label{tab:running_time_20x20}
    \begin{tabular}{c|cc}
        \hline
        \textbf{Agent Number} & \textbf{LLM-NAR} & \textbf{CBS} \\ 
        \hline
        4  & $1.7 \pm 0.5$  & $3.1 \pm 1.3$  \\ 
        8  & $1.9 \pm 0.5$  & $9.3 \pm 3.6$  \\ 
        16 & $2.0 \pm 0.6$  & $32.3 \pm 10.4$ \\ 
        \hline
    \end{tabular}
    \vspace{-0.2in}
\end{table}

\subsection{Real-world Results}
In addition to the simulation study, we further tested the proposed method in real-world experiments using LIMO mobile robots. In the experiments, we obtained the robots' position information at the base station through the Nokov motion capture system. The LIMO robots utilize Mecanum wheel motion, allowing them to perform the up, down, left, and right movements required in MAPF tasks.


In the real-world tests, due to limitations in available space and the number of robots, we evaluated task execution with two to four robots in a $5 \times 4$ size map using LLM-NAR, GPT, and LLaMA3. In the task involving two agents, all three methods successfully completed the task, but the paths generated by LLM-NAR were shorter compared to those produced by GPT and LLaMA3. In the task with four agents, GPT had one robot that failed to reach its target point during execution, and LLaMA3 had two robots fail. In contrast, all robots controlled by LLM-NAR reached their target points, and the paths were shorter. Fig.~\ref{real} shows the trajectories of our LLM-NAR method in the two, three, and four robot tasks.



\section{Conclusion}
In this paper, we addressed the issue of the poor performance of Large Language Models (LLM) in Multi-agent Path Finding (MAPF) by proposing the Neural Algorithmic Reasoners informed Large Language Model for Multi-agent Path Finding (LLM-NAR). We introduced an improved LLM prompting method for MAPF and constructed a Graph Neural Network (GNN)-based Neural Algorithmic Reasoner specifically for MAPF. By integrating the information from both components through a cross-attention mechanism, we developed the LLM-NAR policy. Simulations and real-world experiments demonstrated that our method outperforms other LLM models across various map settings and numbers of agents. In future work, we will further explore utilizing methods such as temporal graph neural networks to optimize the performance of LLM in MAPF tasks.

\bibliographystyle{IEEEtran}
\bibliography{ref}

\end{document}